\documentclass[sigconf]{acmart}

\usepackage{url}            
\usepackage{booktabs}       
\usepackage{amsfonts}       
\usepackage{nicefrac}       
\usepackage{enumitem}
\usepackage{amsmath}
\usepackage{algpseudocode}
\usepackage{longtable}
\usepackage{tcolorbox}
\usepackage{wrapfig}
\usepackage{adjustbox}
\usepackage{booktabs}
\usepackage{tabularx}
\usepackage{float}
\usepackage{marginnote}
\usepackage{booktabs, multirow, array, makecell}
\usepackage[utf8]{inputenc}
\usepackage{booktabs,longtable,array}
\usepackage[dvipsnames]{xcolor}
\usepackage{soul}



\setlist[enumerate]{nosep, left=0pt, labelsep=0.5em, itemsep=0pt, topsep=0pt, parsep=0pt, partopsep=0pt}
\newcolumntype{C}[1]{>{\centering\arraybackslash}m{#1}} 

\setlist[itemize]{nosep, left=0pt, labelsep=0.5em, itemsep=0pt, topsep=0pt, parsep=0pt, partopsep=0pt}
\setlist[enumerate]{nosep, left=0pt, labelsep=0.5em, itemsep=0pt, topsep=0pt, parsep=0pt, partopsep=0pt}
\setlist[description]{nosep, left=0pt, labelsep=0.5em, itemsep=0pt, topsep=0pt, parsep=0pt, partopsep=0pt}

\newcommand{\kk}[1]{{\textcolor{black}{#1}}}
\AtBeginDocument{%
  }

\setcopyright{none}

\acmDOI{}
\acmISBN{}
\acmConference[AI4F @ ICAIF '25]{AI for Finance (AI4F) Workshop at ACM ICAIF 2025}{November 15--18, 2025}{Singapore}


\begin{document}

\title{Synthetic Data-Driven Prompt Tuning for Financial QA over Tables and Documents}

\author{Yaoning Yu}
\affiliation{%
  \institution{University of Illinois Urbana-Champaign}
  \city{Champaign}
  \state{IL}
  \country{United States}
}

\author{Kai-Min Chang}
\affiliation{%
  \institution{U.S. Bank}
  \country{United States}
}

\author{Ye Yu}
\affiliation{%
  \institution{University of Illinois Urbana-Champaign}
  \city{Champaign}
  \state{IL}
  \country{United States}
}

\author{Kai Wei}
\affiliation{%
  \institution{University of South Florida}
  \city{Tampa}
  \state{FL}
  \country{United States}
}

\author{Haojing Luo}
\affiliation{%
  \institution{Starc.institute}
  \country{United States}
}

\author{Haohan Wang}
\affiliation{%
  \institution{University of Illinois Urbana-Champaign}
  \city{Champaign}
  \state{IL}
  \country{United States}
}









\begin{abstract}
\kk{Financial documents like earning reports or balance sheets often involve long tables and multi-page reports. Large language models have become a new tool to help numerical reasoning and understanding these documents. However, prompt quality can have a major effect on how well LLMs perform these financial reasoning tasks.} 
Most current methods tune prompts on fixed datasets of financial text or tabular data, which limits their ability to adapt to new question types or document structures\kk{, or they involve costly and manually labeled/curated dataset to help build the prompts}. We introduce a self-improving prompt framework driven by data-augmented optimization. In this closed-loop process, we generate synthetic financial tables and document excerpts, verify their correctness and robustness, and then update the prompt based on the results. Specifically, our framework combines a synthetic data generator with verifiers and a prompt optimizer, where the generator produces new examples that exposes weaknesses in the current prompt, the verifiers check the validity and robustness of the produced examples, and the optimizer incrementally refines the prompt in response. By \kk{iterating} these steps in a feedback cycle, our method steadily improves prompt accuracy on financial reasoning tasks without needing external labels. Evaluation on DocMath-Eval benchmark demonstrates that our system achieves higher performance in both accuracy and robustness than standard prompt methods, underscoring the value of incorporating synthetic data generation into prompt learning for financial applications.
\end{abstract}

\maketitle

\section{Introduction}
Financial question-answering (QA) involves extracting precise numerical information, such as financial metrics and arithmetic, from earnings reports, balance sheets, and regulatory filings\cite{zhu2021tat, chen2021finqa}. Accurate answers in this domain are critical, as even minor numerical errors or misunderstandings can result in valuation discrepancies amounting to millions of dollars. Despite the importance of precision and reliability, financial QA tasks are typically lengthy, structurally diverse, and heavily rely on tables and multi-page paragraphs, requiring complicated integration of numerical reasoning and textual understanding\cite{chen2022convfinqa, zhao2022multihiertt, li2022learning}. Moreover, much financial data is confidential or propreitary, severly restricting the availablity of extensive public datasets needed for training and fine-tuning sophisticated machine learning models. 



Recent advancements in large language models (LLMs) have shown promising capabilities in numerical reasoning, particularly relevant to financial analysis tasks. LLMs can effectively parse complex tables, handle long-page documents.\cite{phogat2023zero, de2023gpt} However, the accuracy of LLMs depends significantly on carefully crafted prompts. Even minor modifications in prompt wording and formatting can lead to substantial variations in the quality of the model's output, highlighting prompt optimization as a critical challenge in financial document processing by LLMs. In comparison, data augmentation has long been used in supervised learning to improve model robustness and generalization by exposing models to diverse examples of varying difficulties~\cite{mikolajczyk2018data}. In prompt learning, LLMs can produce synthetic financial tables, documents, and numerical queries, thus could avoid related confidential issues. 

Existing approaches for prompt optimization, such as manual tuning, discrete search methods, or gradient-based techniques, have achieved success in certain contexts but typically assume static datasets and a one-time optimization process~\cite{wang2023promptagent, shin2020autopromptelicitingknowledgelanguage, cui2024phaseevo, kwon2024stableprompt, zhang2024revolve}. This assumption limits their ability to adapt and iteratively to emerging financial document formats or evolving question types, restricting their practical effectiveness in dynamic financial environments.

To address these limitations, we propose a self-improving prompting method, a closed-loop prompt optimization framework for numerical reasoning and financial understanding on long tables and documents, through data-augmented optimization. Our system integrates three major components: (1) a Fin-Generator that produces table based and document-based financial queries designed to reveal current prompt weaknesses; (2) Fin-Verifiers ensure the synthetic data's correctness, consistency, and robustness; and (3) a Fin-Prompt Optimizer that uses these examples to refine the prompt iteratively. This feedback loop enables prompts to improve over time by addressing their own failures, without requiring access to new tasks or external supervision.

\textbf{Contributions.} This paper makes the following contributions:
\begin{itemize}
    \item We introduce a self-improving framework for iterative prompt optimization tailored specifically for Finance QA, integrating synthetic financial data generation, verification, and continuous prompt refinement.    
    \item We design a synthetic data generator capable of systematically producing complex and diverse financial queries to proactively identify and address prompt limitations. 
    \item We validate our approach on standard benchmarks for numerical reasoning and financial document understanding, demonstrating consistent improvements over existing prompting methodologies.
\end{itemize}

\section{Related Work}
\subsection{Automatic Prompt Engineering}
Automatically discovering optimal prompts has merged as a crucial area leveraging large language models (LLMs). Automatic Prompt Engineering (APE) techniques typically fall into optimization-based, generative, and template-driven approaches. Optimization-based methods include gradient-based approaches\cite{shin2020autopromptelicitingknowledgelanguage}, reinforcement learning-driven optimization \cite{ouyang2022training, kwon2024stableprompt}, and evolutionary algorithms \cite{cui2024phaseevo}. Generative approaches use LLMs, such as GPT, Claude, and DeepSeek, to produce candidate prompts. For instance, StablePrompt \cite{kwon2024stableprompt} specifically employs reinforcement learning techniques to optimize prompts dynamically. Additionally, PromptAgent \cite{wang2023promptagent} utilizes a sub-goal decomposition strategy, whereas template driven methods use strcutured fill-in-the-blank prompts to enhance clarity and consistency \cite{chen2024promptoptimizationmultisteptasks}.

Recent advancements have expanded these on automatic prompt optimization techniques further. AutoPDL \cite{spiess2025autopdlautomaticpromptoptimization} automates the search for optimal configurations by systematically exploring both agentic and non-agentic prompt patterns using successive halving. The sequential optimal learning approach proposed by \cite{wang2025sequentialoptimallearningapproach} integrates Bayesian regression and Knowledge-Gradient policies for efficient prompt optimization. The Progressively Automatic Prompt Optimization method \cite{qu2025proapoprogressivelyautomaticprompt} introduces an evolutionary algorithm specifically designed to refine prompts for complex visual classification tasks.

In addition, recent directions also address stability, iteration, and efficiency in prompt optimization, including response evolution tracking for stable adjustment \cite{zhang2024revolve}, component-wise refinement of prompt-driven pipelines \cite{xue2025improve}, semantic consistency as a stability signal \cite{chen2025prompt}, and multi-objective prompt compression for efficient reasoning \cite{yu2025premise}.

Our proposed approach integrates hybrid strategies that combine LLM-driven revising, informed by natural language feedback \cite{pryzant2023automaticpromptoptimizationgradient}, along with self-reflection \cite{shinn2024reflexion} and structured planning methodologies \cite{wang2023promptagent}.

\subsection{Data Synthesis}
The use of large language models (LLMs) for data synthesis represents a rapidly evolving frontier, This approach builds upon the foundations established by tradition data augmentation techniques \cite{wang2022toward,liu2023towards}. Recent studies have underscored the ability of LLMs to generate highly fluent, contextually relevant textual data with quality comparable to human-generated content \cite{li2023making, mukherjee2023orca, eldan2023tinystoriessmalllanguagemodels}. For example, Gao et al.\cite{gao2023selfguidednoisefreedatageneration} demonstrated the effectiveness of pre-trained language models (PLMs) in producing task-specific synthetic text beneficial for training and evaluation. Recent work Magpie framework \cite{xu2024magpiealignmentdatasynthesis} capitalizes on the auto-regressive characteristics of aligned LLMs to generate high-quality instructional data. Additionally, synthetic text generation strategy based on gradient matching with human-generated data highlights novel approaches to enhance data quality.\cite{nguyen2025synthetictextgenerationtraining}. Despite these advances, these studies have not fully leveraged advanced methodologies such as chain-of-thought (CoT) reasoning, in-context learning, or data synthesis driven by prompts that integrate task descriptions and label information.

\subsection{Financial Question-Answering}
Financial Question-Answering targets queries over earnings reports, balance sheets, and companies filings, demanding numerical reasoning and financial understanding to interpret tabular data alongside narrative text. Early work emphasized on domain-adaptive pre-training, such as FinBert \cite{araci2019finbert} and BloomberGPT \cite{wu2023bloomberggpt}, to improve coverage of financial terminology and numeric formats. More recent benchmarks-FinQA \cite{chen2021finqa}, ConvFinQA \cite{chen2022convfinqa}, TAT-QA \cite{zhu2021tat}, MultiHiertt \cite{zhao2022multihiertt}, and others\cite{xie2024finben}-evalaute multi-step arithmetic that combines tables values with corresponding passages. To tackle these tasks, existing techniques often employ manually designed zero-shot or few-shot prompts in combination of Program-of-Thought (PoT) and Chain-of-Thought (CoT) prompting\cite{wei2022chain, chen2022program}, while other systems adopt multi-agent architectures with retrieval-augmented generation to enhance LLM's ability in the financial question-answer task\cite{fatemi2024enhancing, wang2025finsage, singh2024finqapt, choi2025finder}.

In this paper, our study experimented with a range of techniques, including in-context learning and prompt-driven data synthesis applied to financial question-answer pairs. Our results indicate that integrating these approaches produces high-quality synthetic financial Q\&A pairs. To further enhance the robustness and applicability of the synthetic data, we introduced a difficulty tier, making the financial questions more challenging. These findings highlight the potential of combining advanced LLM capabilities with tailored prompting strategies to improve data synthesis quality and reliability for financial prompt optimization. 


\section{Method}
\label{sec:methods}

\begin{figure*}
    \centering
    \includegraphics[width=0.8\linewidth]{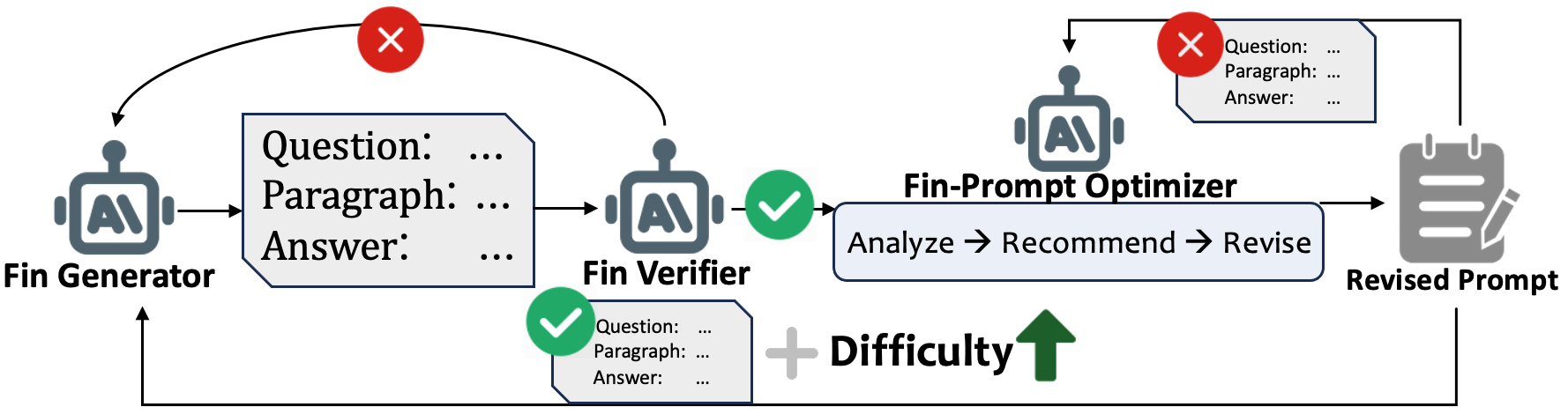}
    \caption{Overview of our proposed iterative prompt optimization workflow. Starting from the left, the Fin-Generator creates a synthetic financial question-answer pair at a specified difficulty level $c$. The generated data then moves to the Fin-Verifiers, which checks its validity and robustness. If the data fails verification, it returns to the Fin-Generator to regenerate a new sample. Once verified, the data proceeds to the Fin-Prompt Optimizer, which systematically improves the current prompt through three sub-steps: analyzing errors, recommending targeted improvements, and revising accordingly. The newly revised prompt is tested against both current and previously resolved cases. If errors persist, the prompt is refined further; if it passes successfully, the system moves on to generate more challenging synthetic data at a higher difficulty level until no error remains.}
    \label{fig:enter-label}
\end{figure*}

Our system presents a multi-agent system for optimizing prompts by implementing data augmentation in numerical financial QA over long tables and multi-page documents. The three module work in cycle: 1) synthetic Fin-Generator creates financial queries and document snippets of increasing difficulty-such as ratio calculations from balance sheets, trend questions on time-series data, and risk-metic lookups in bond prospectuses-to test the current prompt, 2) Fin Verifiers that checks the validity and robustness of generated data, and 3) Fin-Prompt Optimizer iteratively evaluates the model's responses on these synthetic examples, identifies errors, and revise the prompt to improve accuracy. The agents advance in lock-step, refining a concise prompt that adapts to progressively harder samples. An overview of our method is shown in Fig~\ref{fig:enter-label}.

\textbf{\textit{Notation.}}
Let \(\mathcal{X}\) be the space of financial inputs: each \(x\in\mathcal{X}\) is either a table (e.g.\ a row listing revenue, cost, assets) or a document snippet from a bond prospectus. Let \(\mathcal{Y}\subset\mathbb{R}\) denote the set of numeric targets (e.g.\ net income or debt-to-equity ratio). We assume a true joint distribution \(S\) over \(\mathcal{X}\times\mathcal{Y}\). A real dataset of size \(N\) is sampled as
\(\{(x_{i},y_{i})\}_{i=1}^{N}\sim S.\)
We query a fixed LLM with prompt \(p\in\mathcal{P}\); its output is \(f(p,x)\in\mathcal{Y}\). Prediction error is measured by a bounded loss
\(L\bigl(f(p,x),y\bigr)\in[0,1].\)

To enrich prompt optimization, we use a synthetic generator \(q_{\psi}(\tilde x,\tilde y)\) with parameters \(\psi\in\Psi\) to propose \(M\) additional pairs:
\(D=\{(\tilde x_{i},\tilde y_{i})\}_{i=1}^{M}.\)
We estimate an empirical label prior \(p^{*}(y)\) from the real data and regularize the generator by adding
\[
R(\psi)=\mathrm{KL}\bigl(q_{\psi}(y)\,\|\,p^{*}(y)\bigr)
\]
to its objective, penalizing deviation from the true label distribution.

\subsection{Fin-Generator}
The Fin-Generator is responsible for proposing new synthetic financial QA pairs that target shortcomings in the current prompt by sampling latent templates and controlled difficulty tiers.
 
\noindent \textbf{\textit{Sampling rule.}}
First, sample a latent template  
\(
z \;\sim\; g_{\phi}(z \mid S),
\)
and select a difficulty tier \(c\in\{1,\dots,n\}\).  The decoder \(q_{\psi}\) then produces a candidate pair  
\(
(\tilde{x},\tilde{y}) = q_{\psi}(z,\,c),
\)
where \(\tilde{x}\) consists of financial tables and text documents and \(\tilde{y}\) is its numeric answer, designed specifically to probe the prompt’s current weaknesses.  Each candidate will be forwarded to the Fin-Verifiers for validation.  

\noindent \textbf{\textit{Learning objective.}} We fit the parameters \(\psi\) by minimizing a hybrid loss that balances label regularization and prompt-driven error:
\[
\min_{\psi}\;
R(\psi)
\;+\;
\lambda\,
\mathbb{E}_{(\tilde{x},\tilde{y})\sim q_{\psi}}
\bigl[L\bigl(f(p,\tilde{x}),\tilde{y}\bigr)\bigr],
\]
where
\(
R(\psi)
=\mathrm{KL}\bigl(q_{\psi}(y)\,\|\,p^{*}(y)\bigr)
\)
penalizes divergence from the empirical label prior \(p^{*}(y)\) estimated on real financial data.  

\noindent \textbf{\textit{Progressive difficulty.}} To construct a curriculum of rising challenge, sample an ordered sequence of tiers \(c_{1}<\dots<c_{L}\).  Initialize \(z^{(1)}=z\) and generate
\[
(\tilde{x}^{(1)},\tilde{y}^{(1)}) \;=\; q_{\psi}\bigl(z^{(1)},c_{1}\bigr).
\]
For each subsequent level \(k=2,\dots,L\), use a summarizer
\[
z^{(k)} = h_{\phi}\bigl(\tilde{x}^{(k-1)},\tilde{y}^{(k-1)}\bigr)
\]
to distill the previous example into a new latent cue, and then produce
\[
(\tilde{x}^{(k)},\tilde{y}^{(k)}) \;=\; q_{\psi}\bigl(z^{(k)},c_{k}\bigr).
\]
This procedure yields \(L\) examples whose difficulty increases monotonically, providing a structured gradient of challenge for prompt refinement.

\subsection{Fin-Verifiers}

The Fin-Verifiers are implemented as three independent expert voters. Each voter evaluates a synthetic question–answer pair \((\tilde{x},\tilde{y})\) for numerical consistency and structural validity and robustness. A pair is accepted into the synthetic dataset \(D\) only if all three voters approve it. If any voter rejects the pair, it is sent back to the Fin-Generator for regeneration which. This strict consensus rule ensures that only fully validated and robust pairs are retained for prompt optimization.

\subsection{Fin-Prompt Optimizer}

After each synthetic example is generated validly, the Fin-Prompt Optimizer examines the current prompt to uncover any weaknesses and applies targeted revisions before generating the next example. By continuously correcting failures as they arise, the system avoids accumulating hard‐to‐diagnose errors and rapidly converges on a prompt that is robust and fit to the current financial QA task.

\noindent\textbf{\textit{Accuracy.}} At iteration $t\in\{1,\dots,M\}$, we measure how well a prompt $p$ performs on a set $\mathcal{A}\subseteq\mathcal{X}\times\mathcal{Y}$ via
\[
s_{\mathcal{A}}(p)
\;=\;
\frac{1}{|\mathcal{A}|}
\sum_{(\tilde{x},\tilde{y})\in\mathcal{A}}
\mathbb{I}\!\bigl[f(p,\tilde{x})=\tilde{y}\bigr],
\]
where $\mathbb{I}[\cdot]$ is the indicator function (1 if the model’s output matches the target, 0 otherwise).

\medskip
\noindent\textbf{\textit{Step 1: Analyze on failures.}}\label{par:step1}  
Evaluate the current prompt $p^{(t)}$ on the generated synthetic data
\[
D_t \;=\;\{(\tilde{x}_j,\tilde{y}_j)\}_{j=1}^t
\;\subseteq\;\mathcal{X}\times\mathcal{Y},
\]
and collect the current failure
\[
\mathcal{E}^{(t)}
\;=\;
\bigl\{(\tilde{x},\tilde{y})\in D_t
\;\big|\;
f(p^{(t)},\tilde{x})\neq\tilde{y}\bigr\}.
\]
If $\mathcal{E}^{(t)}=\varnothing$, then $p^{(t)}$ already correctly handles every example seen so far and we terminate with $p^*=p^{(t)}$.  Otherwise, proceed to Step 2.

\medskip
\noindent\textbf{\textit{Step 2: Recommend.}}  
A reflection module $\mathcal{R}_{\varphi}$ inspects the current error set and generates a textual patch:
\[
\Delta^{(t)}
\;=\;
\mathcal{R}_{\varphi}\bigl(p^{(t)},\,\mathcal{E}^{(t)}\bigr),
\]
which summarizes why $p^{(t)}$ failed (e.g.\ misunderstood tables and arithmetic errors) and suggests how to amend it.

\medskip
\noindent\textbf{\textit{Step 3: Revise.}}  
A prompt editor $\mathcal{U}_{\theta}$ applies the patch to produce a revised prompt:
\[
\tilde{p}^{(t)}
\;=\;
\mathcal{U}_{\theta}\bigl(\Delta^{(t)},\,p^{(t)},\,\mathcal{E}^{(t)}\bigr).
\]

\noindent\textbf{\textit{Confirmation on Current Error.}}  
Test $\tilde{p}^{(t)}$ on the error slice alone.  If
\(
s_{\mathcal{E}^{(t)}}\bigl(\tilde{p}^{(t)}\bigr)<1,
\)
then some errors remain.  In that case, set $p^{(t)}\leftarrow\tilde{p}^{(t)}$, update $\mathcal{E}^{(t)}$, and repeat Step 2 until this local slice is fully corrected.  Otherwise, move on to global confirmation.

\medskip
\noindent\textbf{\textit{Confirmation on Past Data.}}
Ensure $\tilde{p}^{(t)}$ still succeeds on all previously solved examples by computing
\(
s_{D_t}\bigl(\tilde{p}^{(t)}\bigr).
\)
If new failures appear anywhere in $D_t$, treat them as a fresh error slice and return to Step 2.  If no failures remain, accept $\tilde{p}^{(t)}$ as $p^{(t+1)}$, synthesize the next example, and restart from Step 1 until $t=M$ (or until a user‐specified cap $T_{\max}$).

\medskip
\noindent\textbf{\textit{Convergence guarantee.}}  
Because $s_{D_t}(p^{(t)})$ is non‐decreasing and bounded above by 1, this loop terminates after at most $\min(M,T_{\max})$ successful refinements.  The final prompt
\[
p^*
\;=\;
\arg\max_{0\le t\le T}
s_{D_T}\!\bigl(p^{(t)}\bigr)
\]
achieves perfect coverage ($s_{D_T}(p^*)=1$) whenever that is attainable within budget.

This iterative diagnose-and-repair cycle mirrors how practitioners debug prompts: it isolates specific failures, applies precise fixes, and then verifies that no regressions occur on earlier cases. By repeating this feedback‐driven process, Fin-Prompt Optimizer incrementally constructs prompts that deliver high accuracy, clarity, and generalization in numerical financial QA over tables and documents.

\begin{table*}[!htbp]
\centering
\tiny
\renewcommand{\arraystretch}{1.12}

\resizebox{\textwidth}{!}{%
\begin{tabular}{@{}l c l *{5}{cc}@{}}
\toprule
\multirow{2}{*}{Model} &
\multirow{2}{*}{Size} &
\multirow{2}{*}{Notes} &
\multicolumn{2}{c}{DM$_{\text{SimpShort}}$} &
\multicolumn{2}{c}{DM$_{\text{CompShort}}$} &
\multicolumn{2}{c}{DM$_{\text{SimpLong}}$} &
\multicolumn{2}{c}{DM$_{\text{CompLong}}$} &
\multicolumn{2}{c}{Avg.\ Acc} \\
\cmidrule(lr){4-5}\cmidrule(lr){6-7}\cmidrule(lr){8-9}\cmidrule(lr){10-11}\cmidrule(lr){12-13}
 &  &  & PoT & CoT & PoT & CoT & PoT & CoT & PoT & CoT & PoT & CoT \\
\midrule
\multicolumn{13}{@{}l}{\emph{\textbf{Proprietary LLMs}}}\\
GPT-4o            & --  &      & 84.0 & \textbf{86.0} & 69.5 & 76.5 & 56.0 & \textbf{64.0} & 41.0 & 36.7 & 60.8 & \textbf{62.4}\\
GPT-4-Turbo       & --  &      & 85.5 & 82.5 & 80.0 & \textbf{81.0} & 56.0 & 53.0 & 38.7 & 38.3 & 62.9 & 61.9 \\
Claude-3-Opus     & --  &      & 80.5 & 79.5 & 73.5 & 77.5 & 51.0 & 61.0 & 42.0 & 39.7 & 60.6 & 61.8 \\
Claude-3.5-Sonnet & --  &      & 78.0 & 77.0 & 76.0 & 69.5 & 54.0 & 61.0 & \textbf{44.0} & \textbf{40.0} & 61.8 & 59.2 \\
Claude-3-Sonnet   & --  &      & 82.5 & 80.0 & \textbf{80.5} & 73.0 & 55.0 & 56.0 & 40.3 & 35.3 & 62.7 & 58.5 \\
Gemini-1.5-Flash  & --  &      & 85.0 & 78.0 & 78.5 & 69.5 & 55.0 & 46.0 & 40.0 & 31.7 & 62.8 & 54.5 \\
Gemini-1.5-Pro    & --  &      & 85.5 & 80.5 & 80.0 & 58.0 & \textbf{58.0} & 55.0 & 40.3 & 30.0 & \textbf{63.7} & 52.8 \\
Claude-3-Haiku    & --  &      & 74.5 & 79.0 & 71.5 & 58.5 & 55.0 & 50.0 & 36.7 & 31.7 & 57.1 & 52.5 \\
GPT-4o-Mini       & --  &      & \textbf{88.5} & 69.5 & 77.0 & 69.5 & 53.0 & 56.0 & 38.7 & 28.0 & 62.5 & 52.2 \\
GPT-3.5-Turbo     & --  &      & 71.0 & 60.5 & 52.5 & 39.0 & 41.0 & 28.0 & 28.7 & 15.0 & 46.8 & 34.0 \\
\midrule
\multicolumn{13}{@{}l}{\emph{\textbf{Open-source LLMs}}}\\
DeepSeek-V2  & 236B & MoE  & \textbf{87.0} & 82.0 & 75.5 & 69.5 & \textbf{61.0} & \textbf{56.0} & \textbf{43.0} & \textbf{39.7} & \textbf{64.4} & \textbf{59.8} \\
Mistral-Large & 123B & & 85.0 & \textbf{83.5} & 76.5 & \textbf{81.0} & 56.0 & 55.0 & 41.0 & 31.3 & 62.8 & 59.7 \\
DeepSeek-Coder-V2        & 236B & Code, MoE    & 85.0 & 79.0 & \textbf{78.0} & 66.5 & 56.0 & 54.0 & 41.0 & 37.7 & 63.1 & 57.3 \\
Llama-3.1-70B            &  70B &              & 74.5 & 76.5 & 68.0 & 71.0 & 53.0 & 50.0 & 34.7 & 29.3 & 55.3 & 54.1 \\
Qwen2-72B                &  72B &              & 26.5 & 74.0 & 24.5 & 72.5 &  8.0 & 45.0 &  7.0 & 27.0 & 16.4 & 52.4 \\
Llama-3-70B (Inst)       &  70B &              & 84.5 & 73.5 & 64.0 & 63.5 & 52.0 & 42.0 & 41.0 & 28.3 & 59.0 & 50.1 \\
Mixtral-8x22B            & 141B & MoE          & 30.0 & 74.0 & 21.5 & 57.0 & 25.0 & 47.0 & 14.7 & 24.0 & 21.5 & 47.6 \\
Gemma-2                  &   9B &              & 90.0 & 66.5 & 65.0 & 54.5 & 30.0 & 39.0 & 24.3 & 17.7 & 51.4 & 41.8 \\
DeepSeek-Coder-V2-Lite   &  16B & Code         & 66.0 & 67.5 & 51.0 & 53.5 & 27.0 & 22.0 & 20.3 & 20.9 & 39.0 & 41.6 \\
WizardLM-2               & 141B & MoE          & 62.5 & 60.5 & 56.5 & 55.5 & 25.0 & 34.0 & 17.7 & 18.0 & 39.5 & 39.0 \\
C4AI Command R+          & 104B &              & 35.5 & 65.5 & 39.5 & 51.0 & 19.0 & 31.0 &  8.7 & 18.3 & 24.3 & 39.2 \\
Yi-1.5-9B                &   9B &              & 38.0 & 68.5 & 24.5 & 56.0 & 12.0 & 40.0 & 14.0 & 14.0 & 24.5 & 44.9 \\
Yi-1.5-34B               &  34B &              & 55.4 & 64.5 & 37.5 & 53.0 & 20.0 & 14.0 & 15.3 &  0.0 & 32.4 & 36.9 \\
Mistral-Nemo             &  12B &              & 52.5 & 59.5 & 37.5 & 44.0 & 28.0 & 37.0 & 15.3 & 16.7 & 31.7 & 36.8 \\
Llama-3-8B               &   8B &              & 62.0 & 60.0 & 44.0 & 42.5 & 32.0 & 33.0 & 19.0 & 14.3 & 37.6 & 35.1 \\
DBRX                     & 132B & MoE          & 41.0 & 57.0 & 39.5 & 43.0 & 32.0 & 32.0 & 12.3 & 12.0 & 28.1 & 34.9 \\
Codestoral (CodeLlama)   &  22B & Code         & 39.0 & 51.5 & 38.5 & 41.5 & 18.0 & 23.0 & 17.3 & 17.0 & 28.1 & 31.0 \\
Llama-3-13B              &  13B &              & 36.5 & 56.5 & 21.5 & 31.0 & 24.0 & 29.0 & 12.3 & 17.0 & 28.1 & 31.0 \\
Qwen2-13B                &  13B &              & 33.0 & 56.0 &  9.5 & 33.0 & 14.0 & 31.0 &  2.3 & 10.0 & 24.9 & 29.7 \\
Mathstral                &   7B & Math         & 28.0 & 46.0 & 11.5 & 25.5 &  9.0 & 21.0 &  2.0 & 11.3 & 24.5 & 25.3 \\
GLM-4                    &   9B &              & 69.5 & 44.0 & 53.5 & 34.0 & 33.0 & 33.0 & 22.0 & 17.1 & 41.5 & 25.3 \\
Aya-23                   &   8B &              & 15.1 & 44.0 & 10.5 & 35.0 &  2.0 & 20.0 &  0.0 & 11.7 &  5.6 & 24.3 \\
DeepSeek-V2-Lite         &  16B & MoE          & 70.5 & 45.5 & 18.0 & 17.0 & 10.0 & 10.3 &  3.1 &  0.0 & 21.9 &  0.0 \\
Mixtral-8x7B-v0.1        &  46B & MoE          & 30.0 & 39.0 & 12.0 & 17.0 & 10.0 & 25.0 &  2.1 & 10.6 & 21.9 &  0.0 \\
DeepSeek-Math            &   7B & Math         & 20.5 & 46.0 & 11.0 & 27.0 &  2.0 & 10.0 &  3.0 &  8.0 & 21.8 &  0.0 \\
Llama-2-70B              &  70B &              & 32.5 & 43.5 & 24.0 & 26.0 &  8.0 & 20.0 &  7.0 & 17.1 & 20.8 & 21.9 \\
\midrule
\multicolumn{13}{@{}l}{\emph{\textbf{Our method(Synth. on Short)}}}\\

GPT-4o            & -- & -- 
  & \multicolumn{2}{c}{\underline{\textbf{89}}}
  & \multicolumn{2}{c}{80.5}
  & \multicolumn{2}{c}{\textbf{61}}
  & \multicolumn{2}{c}{\textbf{42}}
  & \multicolumn{2}{c}{\underline{\textbf{68.38}}}\\

GPT-4o-mini      & -- & --   
  & \multicolumn{2}{c}{80.5} 
  & \multicolumn{2}{c}{76}  
  & \multicolumn{2}{c}{51}  
  & \multicolumn{2}{c}{30.33}  
  & \multicolumn{2}{c}{59.46}\\

GPT-4-turbo      & -- & --   
  & \multicolumn{2}{c}{84.5} 
  & \multicolumn{2}{c}{81}  
  & \multicolumn{2}{c}{53}  
  & \multicolumn{2}{c}{41.33}  
  & \multicolumn{2}{c}{\underline{64.96}}\\

Claude-3.5-Sonnet & -- & --  
  & \multicolumn{2}{c}{84} 
  & \multicolumn{2}{c}{\underline{\textbf{81.5}}} 
  & \multicolumn{2}{c}{59}  
  & \multicolumn{2}{c}{41}  
  & \multicolumn{2}{c}{\underline{66.34}}\\
\multicolumn{13}{@{}l}{\emph{\textbf{Our method(Synth. on Long)}}}\\

GPT-4o            & -- & -- 
  & \multicolumn{2}{c}{\textbf{86.5}}
  & \multicolumn{2}{c}{\underline{83}}
  & \multicolumn{2}{c}{\underline{\textbf{66}}}
  & \multicolumn{2}{c}{\textbf{42.67}}
  & \multicolumn{2}{c}{\underline{\textbf{69.54}}}\\

GPT-4o-mini      & -- & --   
  & \multicolumn{2}{c}{80} 
  & \multicolumn{2}{c}{77}  
  & \multicolumn{2}{c}{57}  
  & \multicolumn{2}{c}{34.33}  
  & \multicolumn{2}{c}{62.08}\\

GPT-4-turbo      & -- & --   
  & \multicolumn{2}{c}{84} 
  & \multicolumn{2}{c}{\underline{82.5}}  
  & \multicolumn{2}{c}{57}  
  & \multicolumn{2}{c}{37.33}  
  & \multicolumn{2}{c}{\underline{65.21}}\\

Claude-3.5-Sonnet & -- & --  
  & \multicolumn{2}{c}{84} 
  & \multicolumn{2}{c}{\underline{\textbf{85.5}}} 
  & \multicolumn{2}{c}{61}  
  & \multicolumn{2}{c}{41.33}  
  & \multicolumn{2}{c}{\underline{67.96}}\\
\bottomrule
\end{tabular}%
} 
\caption{Performance on the testmini set of DocMath-Eval across LLMs. Numbers bolded represent the highest performance for each dataset and its method in Propreitary LLMs, Open-source LLMs, and Our method, while numbers underlined represent performance surpasses the original results. For $DM_{CompLong}$, previous methods utilized OpenAI
Embedding 3 Large retriever to retrieve top-10 evidence as input document for downstream evaluations.}
\label{tab:docmatheval}
\end{table*}

\section{Experiments}\label{sec:experiments}
\subsection{Datasets}
We evaluate our approach on the DocMath-Eval benchmark \cite{zhao2023docmath}, which measures numerical reasoning ability within the financial domain.  DocMath-Eval requires models to process lengthy financial documents and perform precise mathematical operations to answer questions. The dataset is divided into four distinct categories:
\begin{enumerate}[label=\arabic*)]
  \item \textbf{Simpshort}: Reannotated from TAT-QA\cite{zhu2021tat} and FinQA\cite{chen2021finqa}, this subset requires simple numerical reasoning over short documents containing a single table. We use all 200 samples from this test-mini set.
  \item \textbf{Simplong}: Reannotated from MultiHiertt\cite{zhao2022multihiertt}, this subset involves straightforward calculations on long documents with multiple tables. We sample all 100 Q\&A pairs from this test-mini set. 
  \item \textbf{Compshort}: Reannotated from TAT-HQA\cite{li2022learning}, this subset demands complex numerical reasoning within short documents featuring a single table. We collect all 200 samples from this test-mini set.
  \item \textbf{Complong}: This subset challenges the model with complex reasoning over long documents that include multiple tables. We select all 300 samples from this test-mini set.
\end{enumerate}

\subsection{Baselines}
We compare our method with existing prompting strategies:

\noindent\textbf{Chain of Thought (CoT)} \cite{wei2022chain} Guides LLM through a sequence of intermediate reasoning steps, enabling it to decompose complex tasks into simpler subproblems.

\noindent\textbf{Program of Thought(PoT)} \cite{chen2022program} Instructs the model to produce a structured program that captures the reasoning process; the final answer is produced by executing this generated program.

\subsection{Implementation}\label{sec:implementation}
\noindent \textbf{Data Generation Prompt Template with Difficulty Levels.} To instantiate our theoretical notion of progressive difficulty, we set each prompt with a maximum level \(c\) and a current level of difficulty with generated synthetic data in the past. From the data generation prompt template below, we demonstrate how to generate a financial question–answer pair from DocMath-Eval by incrementally increasing the complexity of the numerical operations required.

\begin{tcolorbox}[
  colframe=black,
  colbacktitle=black,
  coltitle=white,
  title=Data Generation Prompt Template,
  fonttitle=\bfseries\sffamily\small,
  sharp corners=south,
  boxrule=0.5pt,
  arc=2mm,
  top=2mm,
  bottom=2mm,
  left=2mm,
  right=2mm
]
\small
\texttt{[System Input]}:\\
\\
You are an advanced data-set author who writes realistic, exam-style questions about long financial passages that contain embedded tables and documents. Your task is to generate one DocMath-style question and answer pair. All the samples should be generated differently and the purpose is to challenge the model's ability to reason and answer the question correctly.
.\\[4pt]
\texttt{[User Input]}:\\
\\
\textbf{Guidelines:}
\begin{enumerate}[leftmargin=1.5em,itemsep=0pt]
  \item Make sure that the data generated is different. 
  \item The passage MUST contain a table where each numeric cell is a number.  
  \item The question must require one of(but not limited to, be diverse): a single look-up, a sum, a difference, or a ratio, etc.  
  \item Do not limit by the example, be creative and diverse, and make the question more challenging.
  \item You must generate the answer that is correct based on the question, paragraph, and table.
  \item Generate only one output: the generated content should strictly follow the output format from the examples below.
  \item Difficulty should increase with each iteration, with total difficulty level 15 (current difficulty level: \{c\}).
\end{enumerate}
-----\textit{Continued on Next Page}-----
\end{tcolorbox}

\begin{tcolorbox}[
  colframe=black,
  colbacktitle=black,
  coltitle=white,
  title=Data Generation Prompt Template,
  fonttitle=\bfseries\sffamily\small,
  sharp corners=south,
  boxrule=0.5pt,
  arc=2mm,
  top=2mm,
  bottom=2mm,
  left=2mm,
  right=2mm
]
\small
Past generated samples:
\{Generated data with difficulty\}\\
\\
Below are the Examples with the expected data format:\\
\{True Data 1\}\\
\{True Data 2\}
\end{tcolorbox}

\begin{figure*}
    \centering
    \includegraphics[width=0.95\linewidth]{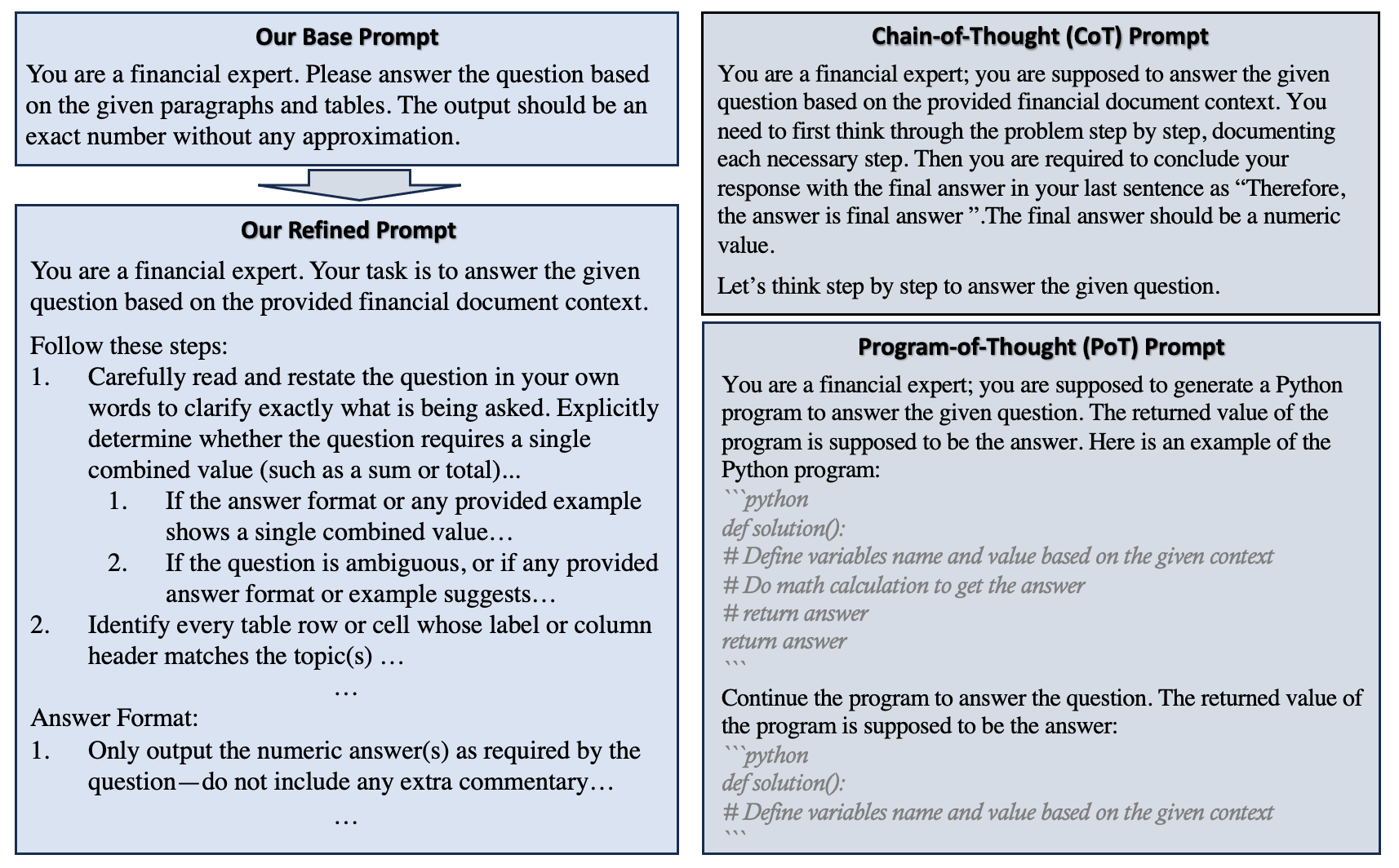}
    \caption{Our approach begins with a base prompt (top-left) and shortened version of its refined prompt synthesized on Long (bottom-left). For comparison, the baseline methods—Chain-of-Thought (top-right) and Program-of-Thought (bottom-right)—are shown on the right side of the figure.}
    \label{fig:prompts}
\end{figure*}

\noindent \textbf{Synthetic Data Generation and Prompt Refinement.}
We developed two prompt templates corresponding to two synthetic data settings. The “Long” template is refined on SimpLong and CompLong synthetic examples—sampling one of each in the prompt, while the “Short” template uses SimpShort and CompShort synthetic samples similarly. By structuring each prompt to include both simple and complex cases, we ensure that our progressive difficulty parameter \(c\) provides a smooth difficulty gain from straightforward lookups to multi‐step reasoning without further subdividing the pipeline to specific datasets (e.g. SimpShort, SimpLong, CompShort, and CompLong). This design guarantees that each prompt learns to handle the full spectrum of financial QA complexity within its designated context length.

We set the maximum difficulty level $c = 15$
for DocMath-Eval and align the number of training iterations $t$ with this difficulty ceiling, such that the optimizer encounters progressively more challenging examples at each step. An example generated by our method at difficulty level 4 (on a 1–10 scale) from the ``Short" subset of DocMath-Eval is shown on the right with additional example in the Appendix~\ref{sec:long_data_generation}. 
The generated data maintains a uniform and structured format, with clear hierarchical segmentation and consistent layout. This formatting rigor ensures both the correctness and stability of the synthesized examples, significantly reducing downstream preprocessing and manual intervention. Furthermore, the structured output improves both interpretability and machine-readability, facilitating smoother analysis workflows and enhancing the reliability and reusability of the synthetic data.

\begin{tcolorbox}[colback=lightgray!10,
    colframe=black,
    arc=2mm,
    boxrule=0.5pt,
    left=2pt,
    right=2pt,
    top=2pt,
    bottom=2pt,
    boxsep=1pt,
    sharp corners,]
\small
\texttt{\textbf{Generated Paragraphs:}\\ 
The company provides a breakdown of expenses over the last four years as follows:\\|Expense Category             |2020|2021|2022|2023|\\|-----------------------------|----|----|----|----|\\|Research \& Development       |320 |350 |370 |400 |\\|Sales \& Marketing            |220 |210 |200 |190 |\\|General \& Administrative     |150 |155 |160 |165 |\\\\ In a separate table, the company also tracks total revenue by region: \\|Region      |2020|2021|2022|2023|\\|------------|----|----|----|----|\\|North America|600|650 |700 |750 |\\|Europe       |300|310 |320 |330 |\\|Asia         |200|220 |240 |260 |\\\\Total expenses and revenues have increased modestly each year, primarily driven by investments in innovation and operational efficiency.}\\
------\textit{Continued on Next Page}------
\end{tcolorbox}

\begin{tcolorbox}[colback=lightgray!10,
    colframe=black,
    arc=2mm,
    boxrule=0.5pt,
    left=2pt,
    right=2pt,
    top=2pt,
    bottom=2pt,
    boxsep=1pt,
    sharp corners,]
\small
\textbf{Generated Question:}\\
What is the combined increase in 'Research \& Development' expense from 2021 to 2023 and 'General \& Administrative' expense from 2020 to 2022 according to the tables? \\
\textbf{Generated Answer:} 60
\end{tcolorbox}



We ran the entire workflow using GPT-4.1 to ensure high-quality synthetic data generation and robust prompt refinement. Specifically, we set the temperature parameter for the Fin-Generator to 0.5, balancing diversity and quality of the generated examples. For the Fin-Verifiers, we set the temperature to 0.0 to rigorously ensure the validity of the synthetic data. In the Fin-Prompt Optimizer stage, we set the temperature to 0.0 during error analysis and revision steps to maximize reliability and stability, while a temperature of 0.5 was used during the recommendation step to encourage diverse and innovative prompt suggestions.

\subsection{Results and Analysis}
\subsubsection{Prompt Refinement}
We present our final refined prompt alongside the base prompt and compare them with CoT and PoT methods in Figure~\ref{fig:prompts} with additional detailed examples provided in Appendix~\ref{sec:refined_prompt}. Our final prompts distinctly emphasize operational level instructions tailored to the specifics of the data. For example, the instruction ``Identify every table row or  cell whose label or column header matches the topic(s) or metric(s) mentioned in the question" directly addresses challenges inherent to financial question-answering tasks involving multiple tables and lengthy documents.

In contrast to baseline methods such as CoT and PoT, which utilize general reasoning instructions not explicitly adapted to the nuances of financial data, our refined prompts automatically guide the model to specifically address the relevant aspects of the given financial context. This targeted instruction enhances the model's capability to accurately handle complex financial numerical reasoning tasks by reducing ambiguity and ensuring consistency in the selection and calculation of relevant data points.

\subsubsection{Evaluations}
We evaluated two specifically designed synthetic-data prompts, “Synthesized on Short” and “Syntheszied on Long”, against traditional Program-of-Thought (PoT) and Chain-of-Thought (CoT) prompts across four datasets in the DocMath-Eval dataset: SimpShort, CompShort, SimpLong, and CompLong. As summarized in Table~\ref{tab:docmatheval}, 
our synthetic-data augmentation approach demonstrates notable improvements, highlighting its effectiveness in enhancing LLM performance on financial question-answering tasks.

\noindent \textbf{SimpShort and CompShort.} In the short-context setting, our ``Synthesized on Short" prompt significantly boosts LLM performance. GPT-4o achieves top results with an accuracy of 89\% on SimpShort and 80.5\% on CompShort, resulting in an overall average accuracy of 68.38\%, which exceeds the best baseline accuracy by 3.98\%. Similarly, Claude-3.5-Sonnet reaches its peak accuracy of 81.5\% on CompShort and 66.34\% on average, surpassing its best baseline by 0.5\% in CompShort and 1.94\% overall.

\noindent \textbf{SimpLong and CompLong.} When extended to the long-context scenario, our ``Synthesized on Long" prompt further enhances results. GPT-4o not only maintains strong performance on shorter tasks (86.5\% SimpShort and 83\% CompShort-again pass the best baseline on CompShort by 2\%) but also significantly improves its performance on longer tasks, achieving 66\% on SimpLong and 42.67\% on CompLong. These outcomes yield a new average accuracy of 69.54\%, surpassing the best baseline by 5.14\%. Claude‑3.5‑Sonnet also achieves a 67.76\% average accuracy and particularly excels on CompShort with an accuracy of 85.5\%, outperforming all other baselines.Even in the highly challenging CompLong scenario
trailing the best baseline by only 1.23\% without relying on any retrieval method.


These results underline the substantial and consistent improvement provided by context-specific synthetic data augmentation, particularly enhancing LLMs' logical and numerical reasoning capabilities for complex financial QA tasks.

\section{Conclusion}
In summary, we propose a novel method that transforms data augmentation into an effective real-time feedback mechanism for enhancing prompt optimization in financial question-answering tasks. Our approach integrates a synthetic data generator, designed to progressively produce more challenging financial queries, with an automated prompt optimizer that iteratively refines prompts based on identified weaknesses. This continuous, feedback-driven process significantly strengthens prompt robustness and accuracy in financial numerical reasoning scenarios.

Empirical results demonstrate that our proposed method consistently outperforms existing prompt-based techniques, achieving substantial accuracy improvements on standard financial reasoning benchmarks. 

By enabling large language models (LLMs) to autonomously identify and rectify prompt's own weaknesses, our framework offers a scalable, adaptive solution that ensures robust generalization and dependable performance on financial QA benchmarks. Future work should adapt more on noisier datasets and complex real-world tasks. This self-improving capability highlights our method's potential for broad application and meaningful collaboration within the financial industry.

\bibliographystyle{ACM-Reference-Format}
\bibliography{papers.bib} 

\onecolumn
\appendix
\section{Generated Data Based from ``Synthesized on Long"}\label{sec:long_data_generation}
We provide an example of generated data in the settings of ``Synthesized on Long" at difficulty level 3 (max difficulty at 15), where the produced synthetic data is learned from Simplong and Complong.
\begin{tcolorbox}[
  colframe=black,
  colbacktitle=black,
  coltitle=white,
  title=Generated Data from SimpLong and CompLong,
  fonttitle=\bfseries\sffamily\small,
  sharp corners=south,
  boxrule=0.5pt,
  arc=2mm,
  top=2mm,
  bottom=2mm,
  left=2mm,
  right=2mm
]
\small
\textbf{Generated Paragraphs:} \\
\\
Part I Item 1 Entergy Corporation, Utility operating companies, and System Energy 253 including the continued effectiveness of the Clean Energy Standards/Zero Emissions Credit program (CES/ZEC), the establishment of certain long-term agreements on acceptable terms with the Energy Research and Development Authority of the State of New York in connection with the CES/ZEC program, and NYPSC approval of the transaction on acceptable terms, Entergy refueled the FitzPatrick plant in January and February 2017. In October 2015, Entergy determined that it would close the Pilgrim plant. The decision came after management\u2019s extensive analysis of the economics and operating life of the plant following the NRC\u2019s decision in September 2015 to place the plant in its \u201cmultiple/repetitive degraded cornerstone column\u201d (Column 4) of its Reactor Oversight Process Action Matrix. The Pilgrim plant is expected to cease operations on May 31, 2019, after refueling in the spring of 2017 and operating through the end of that fuel cycle. In December 2015, Entergy Wholesale Commodities closed on the sale of its 583 MW Rhode Island State Energy Center (RISEC), in Johnston, Rhode Island. The base sales price, excluding adjustments, was approximately \$490 million. Entergy Wholesale Commodities purchased RISEC for \$346 million in December 2011. In December 2016, Entergy announced that it reached an agreement with Consumers Energy to terminate the PPA for the Palisades plant on May 31, 2018. Pursuant to the PPA termination agreement, Consumers Energy will pay Entergy \$172 million for the early termination of the PPA. The PPA termination agreement is subject to regulatory approvals. Separately, and assuming regulatory approvals are obtained for the PPA termination agreement, Entergy intends to shut down the Palisades nuclear power plant permanently on October 1, 2018, after refueling in the spring of 2017 and operating through the end of that fuel cycle. Entergy expects to enter into a new PPA with Consumers Energy under which the plant would continue to operate through October 1, 2018. In January 2017, Entergy announced that it reached a settlement with New York State to shut down Indian Point 2 by April 30, 2020 and Indian Point 3 by April 30, 2021, and resolve all New York State-initiated legal challenges to Indian Point\u2019s operating license renewal. As part of the settlement, New York State has agreed to issue Indian Point\u2019s water quality certification and Coastal Zone Management Act consistency certification and to withdraw its objection to license renewal before the NRC. New York State also has agreed to issue a water discharge permit, which is required regardless of whether the plant is seeking a renewed NRC license. The shutdowns are conditioned, among other things, upon such actions being taken by New York State. Even without opposition, the NRC license renewal process is expected to continue at least into 2018. With the settlement concerning Indian Point, Entergy now has announced plans for the disposition of all of the Entergy Wholesale Commodities nuclear power plants, including the sales of Vermont Yankee and FitzPatrick, and the earlier than previously expected shutdowns of Pilgrim, Palisades, Indian Point 2, and Indian Point 3. See \u201cEntergy Wholesale Commodities Exit from the Merchant Power Business\u201d for further discussion. Property Nuclear Generating Stations Entergy Wholesale Commodities includes the ownership of the following nuclear power plants: |Power Plant|Market|In Service Year|Acquired|Location|Capacity - Reactor Type|License Expiration Date| |Pilgrim (a)|IS0-NE|1972|July 1999|Plymouth, MA|688 MW - Boiling Water|2032 (a)| |FitzPatrick (b)|NYISO|1975|Nov. 2000|Oswego, NY|838 MW - Boiling Water|2034 (b)| |Indian Point 3 (c)|NYISO|1976|Nov. 2000|Buchanan, NY|1,041 MW - Pressurized Water|2015 (c)| |Indian Point 2 (c)|NYISO|1974|Sept. 2001|Buchanan, NY|1,028 MW - Pressurized Water|2013 (c)| |Vermont Yankee (d)|IS0-NE|1972|July 2002|Vernon, VT|605 MW - Boiling Water|2032 (d)| |Palisades (e)|MISO|1971|Apr. 2007|Covert, MI|811 MW - Pressurized Water|2031 (e)| |Consolidated Balance Sheet Data|At July 31,| |(In millions)|2014|2013|2012|2011|2010| |Cash, cash equivalents and investments|\$1,914|\$1,661|\$744|\$1,421|\$1,622| |Long-term investments|31|83|75|63|91| |Working capital|1,200|1,116|258|449|1,074| |Total assets|5,201|5,486|4,684|5,110|5,198| |Current portion of long-term debt|\u2014|\u2014|\u2014|500|\u2014| |Long-term debt|499|499|499|499|998| |Other long-term obligations|203|167|166|175|143| |Total stockholders\u2019 equity|3,078|3,531|2,744|2,616|2,821| ITEM 7 MANAGEMENT\u2019S DISCUSSION AND ANALYSIS OF FINANCIAL CONDITION AND RESULTS OF OPERATIONS Our Management\u2019s Discussion and Analysis of Financial Condition and Results of Operations (MD\&A) includes the following sections: ? Executive Overview that discusses at a high level our operating results and some of the trends that affect our business.  ? Critical Accounting Policies and Estimates that we believe are important to understanding the assumptions and judgments underlying our financial statements.  ? Results of Operations that includes a more detailed discussion of our revenue and expenses.  ? Liquidity and Capital Resources which discusses key aspects of our statements of cash flows, changes in our balance sheets and our financial commitments. You should note that this MD\&A discussion contains forward-looking statements that involve risks and uncertainties. Please see the section entitled \u201cForward-Looking Statements and Risk Factors\u201d at the beginning of Item 1A for important information to consider when evaluating such statements. You should read this MD\&A in conjunction with the financial statements and related notes in Item 8 of this Annual Report. In fiscal 2014 we acquired Check Inc.  and in fiscal 2012 we acquired Demandforce, Inc.  We have included their results of operations in our consolidated results of operations from the dates of acquisition. In fiscal 2013 we completed the sale of our Intuit Websites business and in fiscal 2014 we completed the sales of our Intuit Financial Services (IFS) and Intuit Health businesses. We accounted for all of these businesses as discontinued operations and have therefore reclassified our statements of operations for all periods presented to reflect them as such. We have also reclassified our balance sheets for all periods presented to reflect IFS as discontinued operations. The net assets of Intuit Websites and Intuit Health were not significant, so we have not reclassified our balance sheets for any period presented to reflect them as discontinued operations. Because the cash flows of our Intuit Websites, IFS, and Intuit Health discontinued operations were not material for any period presented, we have not segregated the cash flows of those businesses from continuing operations on our statements of cash flows. See \u201cResults of Operations \u2013 Non-Operating Income and Expense \u2013 Discontinued Operations\u201d later in this Item 7 for more information. Unless otherwise noted, the following discussion pertains to our continuing operations. Executive Overview This overview provides a high level discussion of our operating results and some of the trends that affect our business. We believe that an understanding of these trends is important in order to understand our financial results for fiscal 2014 as well as our future prospects. \\
\textit{continue on next page}
\end{tcolorbox}

\begin{tcolorbox}[colback=lightgray!10,
    colframe=black,
    arc=2mm,
    boxrule=0.5pt,
    left=2pt,
    right=2pt,
    top=2pt,
    bottom=2pt,
    boxsep=1pt,
    sharp corners,]
\small
This summary is not intended to be exhaustive, nor is it intended to be a substitute for the detailed discussion and analysis provided elsewhere in this Annual Report on Form 10-K. See the table later in this Note 7 for more information on the IFS operating results. The carrying amounts of the major classes of assets and liabilities of IFS at July 31, 2013 were as shown in the following table. These carrying amounts approximated fair value. |(In millions)|July 31, 2013| |Accounts receivable|\$40| |Other current assets|4| |Property and equipment, net|31| |Goodwill|914| |Purchased intangible assets, net|4| |Other assets|6| |Total assets|999| |Accounts payable|15| |Accrued compensation|21| |Deferred revenue|3| |Long-term obligations|9| |Total liabilities|48| |Net assets|\$951| Intuit Health In July 2013 management having the authority to do so formally approved a plan to sell our Intuit Health business and on August 19, 2013 we completed the sale for cash consideration that was not significant. We recorded a \$4 million pre-tax loss on the disposal of Intuit Health that was more than offset by a related income tax benefit of approximately \$14 million, resulting in a net gain on disposal of approximately \$10 million in the first quarter of fiscal 2014. The decision to sell the Intuit Health business was a result of management's desire to focus resources on its offerings for small businesses, consumers, and accounting professionals. Intuit Health was part of our former Other Businesses reportable segment. We determined that our Intuit Health business became a long-lived asset held for sale in the fourth quarter of fiscal 2013. A long-lived asset classified as held for sale is measured at the lower of its carrying amount or fair value less cost to sell. Since the carrying value of Intuit Health at July 31, 2013 was less than the estimated fair value less cost to sell, no adjustment to the carrying value of this long-lived asset was necessary at that date. We also classified our Intuit Health business as discontinued operations in the fourth quarter of fiscal 2013 and have segregated its operating results in our statements of operations for all periods presented. See the table later in this Note for more information. We have not segregated the net assets of Intuit Health on our balance sheets for any period presented. Net assets held for sale at July 31, 2013 consisted primarily of operating assets and liabilities that were not material. Because operating cash flows from the Intuit Health business were also not material for any period presented, we have not segregated them from continuing operations on our statements of cash flows. Intuit Websites In July 2012 management having the authority to do so formally approved a plan to sell our Intuit Websites business, which was a component of our Small Business reportable segment. The decision was the result of a shift in our strategy for helping small businesses to establish an online presence. On August 10, 2012 we signed a definitive agreement to sell our Intuit Websites business and on September 17, 2012 we completed the sale for approximately \$60 million in cash. We recorded a gain on disposal of approximately \$32 million, net of income taxes. We determined that our Intuit Websites business became a long-lived asset held for sale in the fourth quarter of fiscal 2012. A long-lived asset classified as held for sale is measured at the lower of its carrying amount or fair value less cost to sell. Since the carrying value of Intuit Websites at July 31, 2012 was less than the estimated fair value less cost to sell, no adjustment to the carrying value of this long-lived asset was necessary at that date.\\
\\
\textbf{Generated Question:} \\
Based on the Consolidated Balance Sheet Data table, what is the sum of the `Long-term investments' for the years 2012 and 2010, and what is the difference between this sum and the `Other long-term obligations' for the year 2013?\\
\\
\textbf{Generated Answer:} 60
\end{tcolorbox}

\newpage
\section{Final Refined Prompts}\label{sec:refined_prompt}
\begin{tcolorbox}[
  colframe=black,
  colbacktitle=black,
  coltitle=white,
  title=Final Refined Prompt Synthesized on Long,
  fonttitle=\bfseries\sffamily\small,
  sharp corners=south,
  boxrule=0.5pt,
  arc=2mm,
  top=2mm,
  bottom=2mm,
  left=2mm,
  right=2mm
]
\small
\texttt{[System Input]}:\\
You are a financial expert. Your task is to answer the given question based on the provided financial document context.\\
\\
Follow these steps:
\begin{enumerate}
  \item Carefully read and restate the question in your own words to clarify exactly what is being asked. Explicitly determine whether the question requires a single combined value (such as a sum or total), multiple separate values, or another specific format. Pay close attention to conjunctions like ``and," ``or,``combined," or similar, and clearly state your interpretation of the question's intent and the expected answer format before proceeding.
  \begin{itemize}
    \item If the answer format or any provided example shows a single combined value (e.g., ``27"), you must sum or otherwise combine the computed values as appropriate and provide only the single combined result, even if the question lists multiple items with ``and."
    \item If the question is ambiguous, or if any provided answer format or example suggests a single combined value, err on the side of providing a single combined value.
  \end{itemize}
  \item If an answer format or example is provided, ensure your output matches that format, combining values if necessary, even if the question lists multiple items with ``and."
  \item Identify every table row or cell whose label or column header matches the topic(s) or metric(s) mentioned in the question.
  \item Extract the exact numeric values that correspond to the requested year(s), quarter(s), or category(ies).
  \item Perform the arithmetic operation(s) described in the question (such as sum, difference, ratio, or combined calculations), ensuring you follow the intent of the question as clarified in step 1. If the question asks for a combined value (for example, the sum of two differences), make sure to provide the single combined result as required.
  \item Double-check your calculations for accuracy and confirm that your answer directly addresses the question as restated, including the expected answer format (single value vs. multiple values).
  \item Never round or approximate the answer; always use the exact result from your calculation.\\
\end{enumerate}

ANSWER FORMAT:
\begin{itemize}
  \item Present your reasoning step by step, documenting each necessary step clearly.
  \item Conclude your response with the final answer in your last sentence, formatted as:
  \item The final answer should be a single numeric value, unless the question specifically requests multiple values. If a single combined value is required, only output that number—do not list intermediate or separate values.
  \item Only output the numeric answer(s) as required by the question—do not include any extra commentary.
\end{itemize}
\end{tcolorbox}

\begin{tcolorbox}[
  colframe=black,
  colbacktitle=black,
  coltitle=white,
  title=Final Refined Prompt Synthesized on Short
  fonttitle=\bfseries\sffamily\small,
  sharp corners=south,
  boxrule=0.5pt,
  arc=2mm,
  top=2mm,
  bottom=2mm,
  left=2mm,
  right=2mm,
]
\small
\texttt{[System Input]}:\\
\\
You are a meticulous financial QA analyst.\\
\\
INPUT FORMAT:  \\
\\
The user message contains two parts separated by a blank line.\\
The first part is the PASSAGE (may include pipe-delimited tables);\\
The second part is the QUESTION.\\
\\
SOLVING STEPS:
\begin{enumerate}
    \item Carefully identify all relevant table rows and columns whose labels, headers, or categories match the topic(s) and entities asked in the question. 
    \item Double-check that the numbers you extract correspond exactly to the correct entity (such as company name), category (such as assets, liabilities), and year or subcategory as specified in the question. 
    \item  Clearly restate which row(s) and column(s) you are using, and echo the exact values you have extracted before performing any calculations.  
    \item Quote the exact wording of the question and explicitly match the order of terms in any arithmetic operation (such as subtraction or ratio), ensuring the calculation reflects the direction or sequence as stated in the question. 
    \item Clearly restate the formula you are using, showing the order of terms as given in the question (e.g., "Net Income - Operating Income" if the question asks for the difference between net income and operating income). 
    \item Perform the arithmetic operation(s) described (sum, difference, ratio, etc.) using the extracted values, and consider whether the result should be positive or negative based on the context and standard financial interpretation.  
    \item Double-check your calculation and ensure all extracted values are from the correct row(s) and column(s).
    \item Never round or approximate the answer when doing calculation, just return the exact result.\\
\end{enumerate}

ANSWER FORMAT:\\
the final numeric answer (never round or approximate the answer when doing calculation, just return the exact result)
\end{tcolorbox}

\end{document}